\pdfoutput=1
\documentclass[runningheads,a4paper]{llncs}
\usepackage[T1]{fontenc}
\usepackage{graphicx}
\usepackage{array}
\usepackage{wrapfig}
\usepackage{placeins}
\usepackage{booktabs}
\usepackage{amsmath}
\usepackage{amssymb}
\usepackage{multirow}
\usepackage{color}
\usepackage{url}

\newcolumntype{L}[1]{>{\raggedright\arraybackslash}p{#1}}
\usepackage[colorlinks=true,urlcolor=blue,citecolor=black,linkcolor=black,breaklinks=true]{hyperref}

\begin{document}

\title{Attention Without Grounding:\\Causal Evaluation of Visual Explanations\\in Medical VLMs}
\titlerunning{Attention Without Grounding in Medical VLMs}

\author{Binesh Sadanandan\and Vahid Behzadan}
\authorrunning{B. Sadanandan and V. Behzadan}
\institute{SAIL Lab, University of New Haven, West Haven, CT, USA\\
\email{\{bsada1@unh,vbehzadan@\}newhaven.edu}}

\maketitle

\begin{abstract}
Attention and saliency heatmaps are increasingly shown as explanations of medical Vision-Language Model (VLM) outputs on chest X-rays (CXR), yet whether a heatmap points to the image evidence that actually drives the answer has not been tested causally. We audit explanation faithfulness along three axes: overlap with radiologist bounding boxes on PadChest~\cite{bustos2020padchest} ($n{=}637$), attribution mass inside radiologist pixel masks on CheXlocalize~\cite{saporta2022benchmarking} ($n{=}643$), and $16{\times}16$ patch-occlusion causal maps that record which regions, when hidden, change the prediction. We evaluate three MedGemma-4B variants~\cite{sellergren2025medgemma,anon_psflora2026} with cross-family probes on LLaVA-RAD~\cite{llavarad2024}, Qwen3-VL-8B-Instruct~\cite{qwen3vl2025}, and the specialist CheXagent-2-3b~\cite{chen2024chexagent}; two CXR-trained classifiers (DenseNet121, ResNet50)~\cite{selvaraju2017gradcam,cohen2022torchxrayvision} are positive controls.
\textbf{A faithful heatmap needs two things at once: the model must use the image, and its attention must fall on the regions whose occlusion changes the answer. No evaluated VLM satisfies both.} The MedGemma variants and the Qwen3-VL probe use the image, but their attention anti-correlates with patch-occlusion causal importance ($-$0.098 base, $-$0.050 targeted LoRA, $-$0.168 full LoRA, $-$0.117 Qwen3-VL; all 95\% bootstrap confidence intervals below zero). LLaVA-RAD's attention correlates positively, but only because it barely uses the image (99.1\% text-only agreement, near-zero causal mass), so its $\rho$ joins two near-zero signals. Attention also misses the annotated anatomy: true-region overlap never beats shifted or random controls, and no method places more than 22\% of its mass inside the radiologist mask. Two CXR-trained classifiers pass the same protocol ($p{<}10^{-4}$ over shuffled on every metric), so the failure is specific to the heatmaps, not the metric. These heatmaps are reassuring but not faithful: clinical explanations need controlled localization metrics and causal perturbation, not visual inspection alone.
\end{abstract}

\section{Introduction}\label{sec:intro}

When a radiology VLM says ``cardiomegaly present,'' the natural follow-up is: what part of the image drove that answer? The usual reply is an attention heatmap. If the hot region lands on the heart silhouette, the explanation feels reassuring. But reassurance is not evidence. We ask whether these heatmaps point to clinically relevant image evidence, and what causal probe can corroborate that judgment, an interpretability question that visual inspection alone cannot settle.

The faithfulness of attention as explanation is contested in language models~\cite{jain2019attention,wiegreffe2019attention} and limited for chest-X-ray classifiers~\cite{saporta2022benchmarking}. Whether VLMs, which jointly attend over text and image tokens with generative decoding, produce more faithful explanations is unresolved; we extend the evaluation to multiple attribution methods, controlled bounding-box and pixel-mask baselines, and patch-level causal perturbation.

We study MedGemma-4B~\cite{sellergren2025medgemma} in three configurations: the base model, a targeted Low-Rank Adaptation (LoRA) on layers 15\textendash 19~\cite{anon_psflora2026}, and a full LoRA across all 34 layers. Three cross-family probes test whether the behaviour is architecture-specific. LLaVA-RAD~\cite{llavarad2024} is a 7B Vicuna model with a BiomedCLIP-CXR encoder; Qwen3-VL-8B-Instruct~\cite{qwen3vl2025} is a general-purpose VLM; and CheXagent-2-3b~\cite{chen2024chexagent} is a chest-X-ray specialist. Two CXR-trained classifiers, DenseNet121-Chex and ResNet50-All, provide positive localization controls through Grad-CAM~\cite{selvaraju2017gradcam,cohen2022torchxrayvision}. These families span a wide range of vision encoders and decoders, so a consistent failure cannot be pinned on one architecture. We evaluate on MIMIC-CXR~\cite{johnson2019mimic}, on PadChest~\cite{bustos2020padchest} with radiologist bounding boxes for $n{=}637$ cases, and on CheXlocalize~\cite{saporta2022benchmarking} with pixel masks across ten pathologies for $n{=}643$ cases.

\textbf{Contributions.} We make three claims, each established by causal perturbation rather than by visual inspection. \emph{(i)}~In every evaluated medical VLM, attention and saliency heatmaps do not track causal importance: true-region overlap does not exceed random controls, and the attention-causal Spearman $\rho$ is negative for the MedGemma variants and Qwen3-VL, with LLaVA-RAD reversing sign only because its causal mass is near zero. \emph{(ii)}~The failure is not an artifact of method, modality, or metric: it holds across six attribution methods, including the axiomatic integrated gradients~\cite{sundararajan2017axiomatic} and the model-agnostic RISE~\cite{petsiuk2018rise}, while the same protocol detects strong localization in the CXR-trained classifier baselines. \emph{(iii)}~Grounding is achievable but not by these general medical VLMs: a CXR specialist is causally grounded on the annotation style it matches yet not on a second dataset, and fine-tuning that reduces paraphrase flips does not improve grounding.

\textbf{What we do not claim.} We do not claim that all VLM explanations fail, that the classifier baselines are clinically superior, or that we have exhausted the attribution-method space. The practical claim is narrow and falsifiable: visual inspection of these heatmaps is not a reliable test of grounding.

\section{Related Work}\label{sec:related}

Attention faithfulness is contested in NLP~\cite{jain2019attention,wiegreffe2019attention} and benchmarked for chest-X-ray classifiers~\cite{saporta2022benchmarking,adebayo2018sanity,han2023generalist}. Medical-VLM trustworthiness benchmarks measure paraphrase sensitivity~\cite{anon_psfmed2026,ribeiro2020checklist}, modality bias~\cite{decoct2025}, hallucination~\cite{medvh2025,xia2024cares}, and uncertainty under shift~\cite{anon_uq2026} individually. Concurrent work finds that vision-language benchmarks barely test visual grounding~\cite{lan2026seeing}, that medical VLMs trade grounding for sycophantic agreement under social pressure~\cite{aranya2026sycophancy}, and that they default to anatomical priors on rare findings~\cite{mayer2025sixfingers}; we add a direct causal-faithfulness audit of the heatmaps these models produce. Mechanistic work~\cite{anon_psflora2026} traces paraphrase consistency to a layer-17 sparse autoencoder feature and reduces flips via targeted LoRA; we reuse that checkpoint and contribute the grounding, multi-method attribution, and counterfactual analyses.

\section{Methods}\label{sec:methods}

\begin{figure}[t]
\centering
\includegraphics[width=\textwidth]{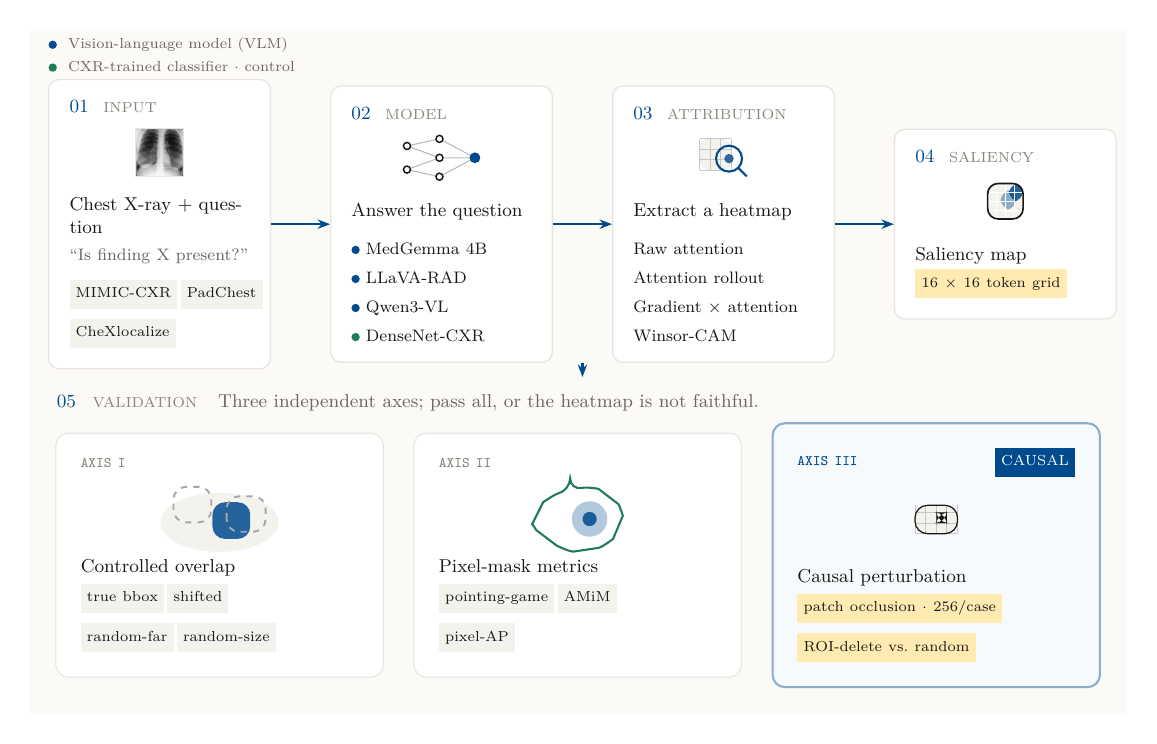}
\caption{Explanation-validation protocol. A saliency map (pipeline, top) is tested along three independent axes: (i) controlled overlap against shifted, random-far, and random-size bounding boxes on PadChest; (ii) pixel-mask metrics (pointing-game, attribution mass in mask, pixel-AP) against radiologist segmentations on CheXlocalize; (iii) causal perturbation via $16{\times}16$ patch occlusion (256 forward passes per case). A heatmap is faithful only if overlap exceeds the controls \emph{and} high-saliency patches are causally important; in this audit every evaluated VLM fails at least one axis, while the CXR-trained classifier controls pass all three.}
\label{fig:workflow}
\end{figure}

\subsection{Models and Datasets}

We evaluate eight configurations across six model families (Table~\ref{tab:models}): MedGemma-4B-IT~\cite{sellergren2025medgemma} in three configurations, two cross-family VLM probes, a chest-X-ray specialist VLM, and two CXR-trained classifier baselines.

\begin{table}[t]
\centering
\caption{Models evaluated. LoRA adapters use rank 16, $\alpha{=}32$, dropout 0.05, lr $2{\times}10^{-4}$, batch 8, at $n{=}500$ binary-presence and $n{=}12$K multi-task scales; the 98 presence test items are held out from both. Classifiers use class-specific Grad-CAM~\cite{selvaraju2017gradcam} upsampled to $224{\times}224$.}
\label{tab:models}
{\scriptsize
\setlength{\tabcolsep}{2.5pt}
\begin{tabular}{@{}L{0.27\textwidth}L{0.58\textwidth}L{0.08\textwidth}@{}}
\toprule
Model & Backbone / configuration & Role \\
\midrule
MedGemma-4B Base & Gemma~2 + SigLIP; 256 tokens ($16{\times}16$) & VLM \\
\quad Targeted LoRA & rank-16 Q/K/V, L15--19; SAE feature 3818 @ L17~\cite{lieberum2024gemma} & tuned \\
\quad Full LoRA & rank-16 all projections, L0--33 & tuned \\
LLaVA-RAD-7B~\cite{llavarad2024} & Vicuna-7B + BiomedCLIP-CXR & probe \\
Qwen3-VL-8B-\allowbreak Instruct~\cite{qwen3vl2025} & general-purpose dense VLM ($8{\times}8$ grid) & probe \\
CheXagent-2-3b~\cite{chen2024chexagent} & chest-X-ray specialist VLM & decision \\
DenseNet121-Chex~\cite{cohen2022torchxrayvision} & \texttt{densenet121-res224-chex} @ \texttt{features.norm5} & control \\
ResNet50-All~\cite{cohen2022torchxrayvision} & \texttt{resnet50-res512-all} (multi-dataset) @ \texttt{model.layer4} & control \\
\bottomrule
\end{tabular}}
\end{table}

\textbf{Datasets.} MIMIC-CXR~\cite{johnson2019mimic} (98 binary presence questions, 3\textendash 5 paraphrases each, credentialed); PadChest~\cite{bustos2020padchest} (861 binary questions, 637 with radiologist bounding boxes across eight finding groups, public); CheXlocalize~\cite{saporta2022benchmarking} validation split (643 samples with pixel masks across ten pathologies, public); and VinDr-CXR~\cite{nguyen2022vindr} ($n{=}300$ presence questions with radiologist boxes) as a secondary replication set. Paraphrases come from template-based rules over five linguistic phenomena, filtered with BioClinicalBERT~\cite{alsentzer2019bioclinical} cosine similarity $>0.95$; no LLM-generated text is used.

\textbf{Reproducibility.} MedGemma-4B-IT~\cite{sellergren2025medgemma}, LLaVA-RAD~\cite{llavarad2024}, Qwen3-VL-8B-Instruct~\cite{qwen3vl2025}, and the DenseNet-CXR weights via TorchXRayVision~\cite{cohen2022torchxrayvision} are available under their licenses; LoRA follows Hu et al.~\cite{hu2022lora} defaults with layer selection from~\cite{anon_psflora2026}. All metrics report 95\% bootstrap CIs (2000 resamples). Per-pathology paired permutation tests (true vs.\ shuffled saliency, 10{,}000 resamples) are Holm-Bonferroni corrected across the ten CheXlocalize pathologies at $\alpha{=}0.05$; pooled paired Cohen's $d$ quantifies effect size. Even the strongest VLM configuration (full LoRA, gradient-attention) is heterogeneous: 8 of 10 pathologies are Holm-significant on at least one metric, the only exceptions being the two lowest-prevalence classes (Pneumothorax $n{=}8$, Lung Lesion $n{=}1$); the well-powered classes ($n{=}33$--$125$) drive the aggregate. Code and the evaluation protocol are at \url{https://github.com/thedatasense/medicalvlm_attention_without_grounding}; the $n{=}12$K LoRA adapters are a Hugging Face collection.\footnote{\url{https://huggingface.co/collections/saillab/mechanistically-guided-lora-chil-2026-69ff7afccce7547a00180b2a}} CheXagent-2-3b requires a separate Transformers~4.40 / float32 environment.

\subsection{Evaluation Protocol}\label{sec:protocol}

The protocol (Fig.~\ref{fig:workflow}) has three axes. \textbf{Grounding overlap:} attention from the last text token to the 256 image tokens, averaged over heads and layers 10\textendash 20 and thresholded at the 90th percentile, scored against the bounding-box mask versus true, shifted, random-far, and random-size controls (with a nine-group layer sweep). \textbf{Pixel-mask attribution:} four methods (raw attention, attention rollout~\cite{abnar2020quantifying}, gradient-attention product~\cite{chefer2021transformer}, Winsor-CAM), upsampled to $224{\times}224$ and scored against radiologist masks by pointing-game hit rate, attribution mass in mask (AMiM), and pixel-AP with paired permutation. \textbf{Counterfactual perturbation:} patch occlusion mean-fills each of the 256 patches and records the margin shift $\delta_{r,c}=m_{\text{orig}}-m_{\text{occluded}}$ (positive when occluding a patch reduces the answer margin) to build a $16{\times}16$ causal map; the main table correlates against the signed shift, the per-method comparison against $|\delta_{r,c}|$. We add integrated gradients~\cite{sundararajan2017axiomatic} and RISE~\cite{petsiuk2018rise} on this axis (six methods total) and report insertion/deletion AUC. Mid-gray fill is used throughout, reported as a known protocol choice.

\section{Experiments and Results}\label{sec:results}

\textbf{Behavioural context.} Companion work~\cite{anon_psfmed2026,anon_psflora2026} shows that reduced paraphrase sensitivity coincides with reduced image use (LLaVA-RAD: 99.1\% text-only agreement; Full LoRA: up to 77.9\%), motivating whether these heatmaps are grounded.

\subsection{Attention Grounding Failure}\label{sec:results_grounding}

\textbf{Bounding-box overlap.} For the base MedGemma model on PadChest ($n{=}637$), true-box attention coverage (0.037, 95\% CI 0.031--0.044) is the \emph{lowest} of four conditions, below shifted (0.042), random far (0.052), and random sized (0.048) boxes. The pattern holds for both LoRA variants and across a nine-group layer sweep, so it is not a layer-selection artifact. LLaVA-RAD is weaker still (true-box coverage 1.2\%).

\begin{figure}[t]
\centering
\includegraphics[width=\textwidth]{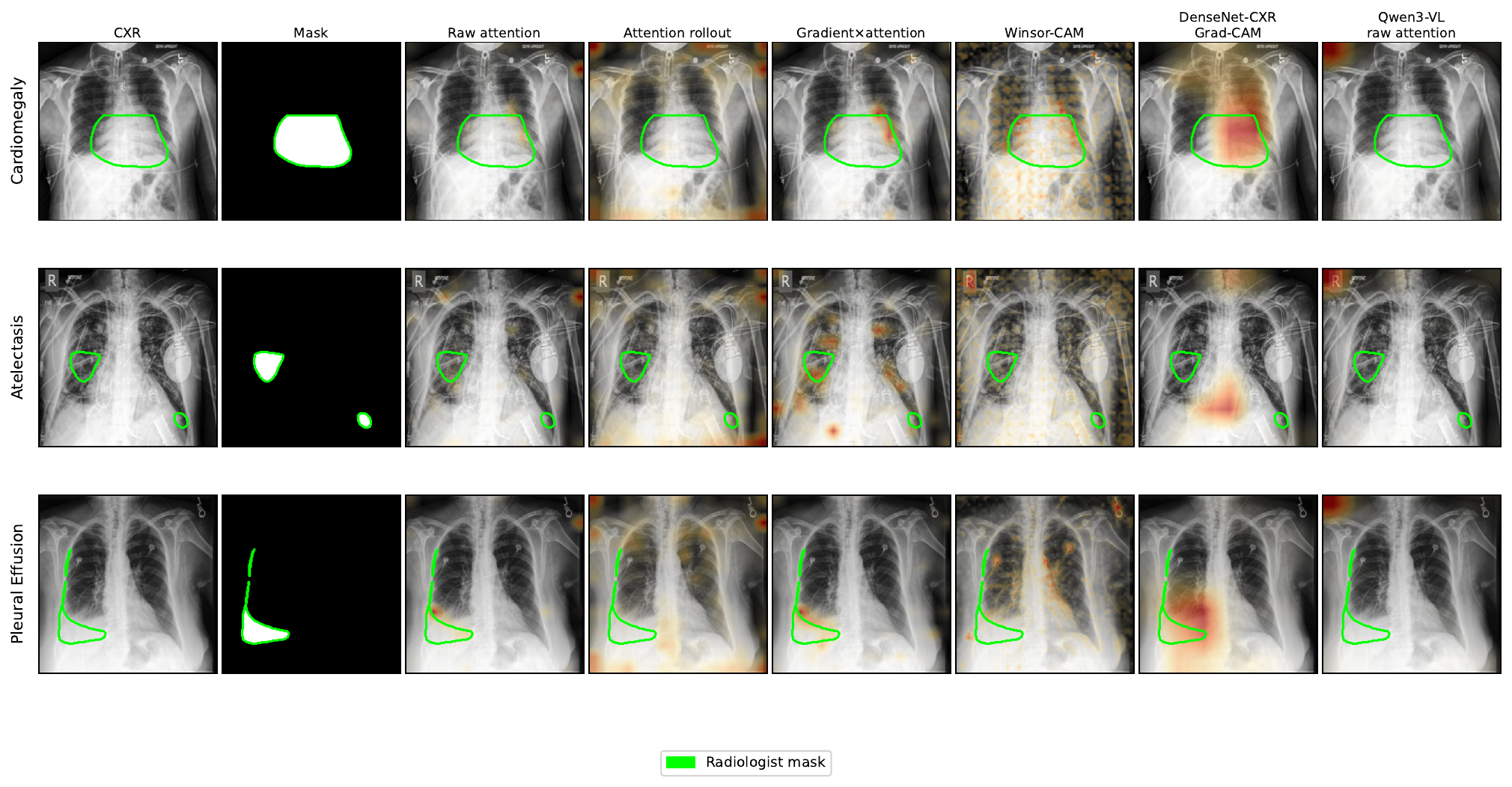}
\caption{Qualitative attribution comparison on three CheXlocalize cases. Columns show the CXR, radiologist mask, MedGemma-4B base methods, DenseNet-CXR Grad-CAM, and Qwen3-VL raw attention. MedGemma maps are broad or near-empty and Qwen3-VL is nearly uniform; DenseNet-CXR is the only method whose peak reliably falls near the mask.}
\label{fig:qualitative}
\end{figure}

\textbf{Multi-method evaluation on pixel masks (Table~\ref{tab:attribution_methods}; Fig.~\ref{fig:qualitative}).}
On CheXlocalize~\cite{saporta2022benchmarking} pixel masks ($n{=}643$), no VLM attribution method places more than 22\% of saliency mass inside the radiologist mask. Gradient-attention product reaches the highest VLM pointing-game (44--51\%) but AMiM stays $\leq 0.20$; raw attention improves with fine-tuning (PG $7.2\%\to 42.9\%$) but AMiM saturates at 0.214; rollout and Winsor-CAM sit at or below the random-mask baseline, and the Qwen3-VL probe is weakest (PG exactly 0\%). \emph{In contrast}, both CXR-trained classifiers produce $p{<}10^{-4}$ gaps over shuffled on every metric, so the negative VLM result is not an artifact of the imaging modality or the localization metric.

\begin{table}[t]
\centering
\caption{Attribution method comparison on CheXlocalize ($n{=}643$). PG: pointing-game hit rate (\%). AMiM: attribution mass in mask. Pixel-AP: pixel average precision. $\Delta$: gap between true and shuffled-mask saliency (positive favors true). Best VLM $\Delta$ per metric is \textbf{bold}. Among VLMs no method exceeds 22\% AMiM; rollout and Winsor-CAM give $\Delta{\approx}0$. Both CXR classifiers produce larger gaps on the same images and classes ($p{<}10^{-4}$).}
\label{tab:attribution_methods}
{\scriptsize
\begin{tabular}{@{}llcccccc@{}}
\toprule
& & \multicolumn{2}{c}{PG (\%)} & \multicolumn{2}{c}{AMiM} & \multicolumn{2}{c}{Pixel-AP} \\
\cmidrule(lr){3-4}\cmidrule(lr){5-6}\cmidrule(lr){7-8}
Method & Model & true & $\Delta$ & true & $\Delta$ & true & $\Delta$ \\
\midrule
  Raw attention & Base & 7.2 & +3.9 & 0.158 & +0.043 & 0.211 & +0.073 \\
   & Targeted LoRA & 21.0 & +12.0 & 0.189 & +0.059 & 0.264 & +0.107 \\
   & Full LoRA & 42.9 & +22.6 & 0.214 & \textbf{+0.066} & 0.340 & \textbf{+0.149} \\
\midrule
  Attention rollout & all 3 MedGemma & 0.3 & +0.2 & 0.073 & $\approx 0$ & 0.068 & $\approx 0$ \\
\midrule
  Gradient attention & Base & 44.3 & +20.8 & 0.178 & +0.044 & 0.296 & +0.113 \\
   & Targeted LoRA & 39.7 & +17.1 & 0.176 & +0.044 & 0.279 & +0.103 \\
   & Full LoRA & 50.9 & \textbf{+26.4} & 0.200 & +0.057 & 0.333 & +0.139 \\
\midrule
  Winsor-CAM & Base & 12.0 & +2.2 & 0.109 & +0.004 & 0.143 & +0.018 \\
   & Targeted LoRA & 7.9 & $-$0.2 & 0.099 & +0.002 & 0.120 & +0.010 \\
   & Full LoRA & 9.8 & +3.1 & 0.103 & +0.006 & 0.126 & +0.015 \\
\midrule
\multicolumn{8}{@{}l}{\textit{Cross-architecture probes:}} \\
  Raw attention & LLaVA-RAD & 0.2 & $-$0.2 & 0.076 & +0.015 & 0.103 & +0.017 \\
  Raw attention & Qwen3-VL-8B & 0.0 & +0.0 & 0.063 & +0.005 & 0.081 & +0.006 \\
\midrule
\multicolumn{8}{@{}l}{\textit{Localization-only baselines (CXR CNN classifiers, $n{=}536$ matched findings):}} \\
  Grad-CAM & DenseNet-CXR-Chex & 37.5 & +12.5 & 0.209 & +0.046 & 0.306 & +0.074 \\
  Grad-CAM & ResNet50-CXR-All & 28.7 & +13.6 & 0.170 & +0.043 & 0.213 & +0.060 \\
\bottomrule
\end{tabular}}
\end{table}

\subsection{Counterfactual Perturbations}\label{sec:results_counterfactual}

Table~\ref{tab:counterfactual} summarises patch-occlusion causal grounding across the MedGemma variants and cross-family probes on PadChest. Spearman $\rho$ between raw-attention saliency and the causal-importance map is negative for every MedGemma variant ($-$0.098 base, $-$0.050 targeted LoRA, $-$0.168 full LoRA at $n{=}500$), persisting at the $n{=}12$K multi-task scale-up. Qwen3-VL agrees in direction ($\rho{=}{-}0.117$ on its native $8{\times}8$ grid, $n{=}637$). LLaVA-RAD reverses sign ($\rho{=}{+}0.137$), but its causal AMiM is close to zero (0.005 vs.\ 0.066 attention): it barely uses the image, so its attention agrees with the very small causal signal that exists, not clinical anatomy. For the three MedGemma variants the AMiM difference (causal $-$ attention) is positive and significant ($p_{\text{perm}}{<}0.001$, $n{=}474$), largest for full LoRA, the variant with the lowest flip rate: consistency-improving fine-tuning does not align attention with cause. Restricting to the $n{=}474$ images all five models processed preserves the pattern, with all 95\% CIs excluding zero.

The mismatch is not specific to raw attention (Table~\ref{tab:counterfactual}, lower panel): rollout, gradient$\times$attention, and the axiomatic integrated gradients~\cite{sundararajan2017axiomatic} all anti-correlate with causal importance for every MedGemma variant ($-$0.18 to $-$0.28 against the absolute-shift causal map, $n{=}474$, CIs exclude zero), Winsor-CAM is near zero, and the model-agnostic RISE~\cite{petsiuk2018rise} is near-uniform ($\rho$ between $+$0.009 and $+$0.010, stable to $N{=}1500$ masks) because the models barely respond to random masking.

\textbf{Second annotated dataset (VinDr-CXR).} The grounding failure replicates on VinDr-CXR~\cite{nguyen2022vindr} ($n{=}300$ presence questions with radiologist boxes per model). Attention mass in the radiologist box stays low (0.075--0.105), below the causal map (0.100--0.133), and true-box AMiM exceeds shuffled controls on only 27--33\% of cases, so attention again fails to concentrate on annotated anatomy. The attention-causal correlation is near zero ($\rho{=}{+}0.061$ base, $+$0.061 targeted LoRA, $+$0.053 full LoRA), so attention does not track causal importance on a second dataset either; the residual correlation is dataset-dependent in sign ($|\rho|<0.17$ throughout), itself a reason not to read heatmap-causal alignment as reliable.

\textbf{CXR-specialist VLM (CheXagent): grounding is dataset-dependent.} CheXagent-2-3b~\cite{chen2024chexagent} does not expose cross-modal attention, so we probe its decision with the margin-only ROI test (occlude the radiologist box vs.\ a random same-size region). It is not grounded on PadChest ($+$0.038, 95\% CI $[-0.025, +0.100]$) but \emph{is} on VinDr-CXR ($+$0.568 gray, $+$0.692 blur); the effect strengthens under blur, ruling out a gray-opacity artifact. A specialist VLM can be causally grounded on the annotation style it matches but not in general.

\textbf{Insertion/deletion fidelity.}
On the standard insertion/deletion protocol ($n{=}200$ per VLM), DenseNet-CXR shows a real saliency effect (insertion AUC 0.690 vs.\ random 0.666, $\Delta{=}{+}0.024$, $p_{\text{perm}}{<}10^{-4}$) while all five VLMs have uniformly small effects ($|\Delta\text{AUC}|\leq 0.024$). The VLM methods fail all three probes (annotation alignment, input-attribution alignment, the cross-architecture sanity check) in the same direction while DenseNet passes the annotation and insertion probes, so the result holds across independent axes rather than one metric.

\begin{table}[t]
\centering
\caption{Patch-occlusion causal grounding on PadChest. \emph{Top:} raw-attention Spearman $\rho$ vs.\ the signed causal map; AMiM $\Delta$ is causal minus attention mass inside the radiologist bbox ($p_{\text{perm}}{<}0.001$ where shown; 95\% CIs for $\rho$). \emph{Bottom:} per-method $\rho$ vs.\ absolute margin shift, explaining why raw attention differs from the signed-$\rho$ value. Negative $\rho$ persists under the $24{\times}$ scale-up.}
\label{tab:counterfactual}
{\scriptsize
\setlength{\tabcolsep}{2.4pt}
\begin{tabular}{@{}lcccccc@{}}
\toprule
Model & $\rho$ & 95\% CI & $L_1$ & Causal & Attn. & $\Delta$ \\
\midrule
\multicolumn{7}{@{}l}{\textit{MedGemma family ($n{=}474$; $16{\times}16$ grid):}} \\
Base & $-$0.098 & [$-$0.121,$-$0.075] & 1.440 & 0.148 & 0.087 & $+$0.061 \\
Targeted LoRA (500) & $-$0.050 & [$-$0.071,$-$0.030] & 1.442 & 0.147 & 0.085 & $+$0.062 \\
Targeted LoRA (12K MT) & $-$0.115 & [$-$0.136,$-$0.095] & 1.388 & 0.160 & 0.092 & $+$0.068 \\
Full LoRA (500) & $-$0.168 & [$-$0.190,$-$0.146] & 1.433 & 0.180 & 0.091 & $+$0.088 \\
Full LoRA (12K MT) & $-$0.145 & [$-$0.167,$-$0.123] & 1.443 & 0.162 & 0.085 & $+$0.077 \\
\midrule
\multicolumn{7}{@{}l}{\textit{Cross-family probes ($n{=}637$; native grid):}} \\
Qwen3-VL-8B ($8{\times}8$) & $-$0.117 & [$-$0.130,$-$0.104] & 1.481 & 0.031 & 0.198 & $-$0.167 \\
LLaVA-RAD ($16{\times}16$) & $+$0.137 & [$+$0.117,$+$0.156] & 1.276 & 0.005 & 0.066 & $-$0.062 \\
\bottomrule
\end{tabular}

\smallskip
\textit{Per-method $\rho$ vs.\ the absolute-shift causal map ($n{=}474$; RISE $n{=}200$, $N{=}320$):}\\[2pt]
\begin{tabular}{@{}lccc@{}}
\toprule
Method & Base & Targeted LoRA & Full LoRA \\
\midrule
Raw attention & $-$0.263 & $-$0.202 & $-$0.253 \\
Attention rollout & $-$0.257 & $-$0.178 & $-$0.238 \\
Gradient $\times$ attention & $-$0.243 & $-$0.189 & $-$0.258 \\
Winsor-CAM & $-$0.059 & $+$0.052 & $+$0.040 \\
Integrated gradients & $-$0.276 & $-$0.208 & $-$0.281 \\
RISE & $+$0.010 & $+$0.009 & $+$0.009 \\
\bottomrule
\end{tabular}}
\end{table}

\FloatBarrier
\section{Discussion and Conclusion}\label{sec:discussion}

\textbf{Consistency without grounding.}
Reducing paraphrase flip rates does not produce models that look at the right anatomy. Fine-tuning trades consistency for text-only agreement and yields only modest grounding gains: raw-attention AMiM still leaves four fifths of mass outside the mask, and Winsor-CAM does not improve. Consistency tells you about the mechanism, not whether the model reads the evidence.

\begin{figure}[t]
\centering
\includegraphics[width=0.60\textwidth]{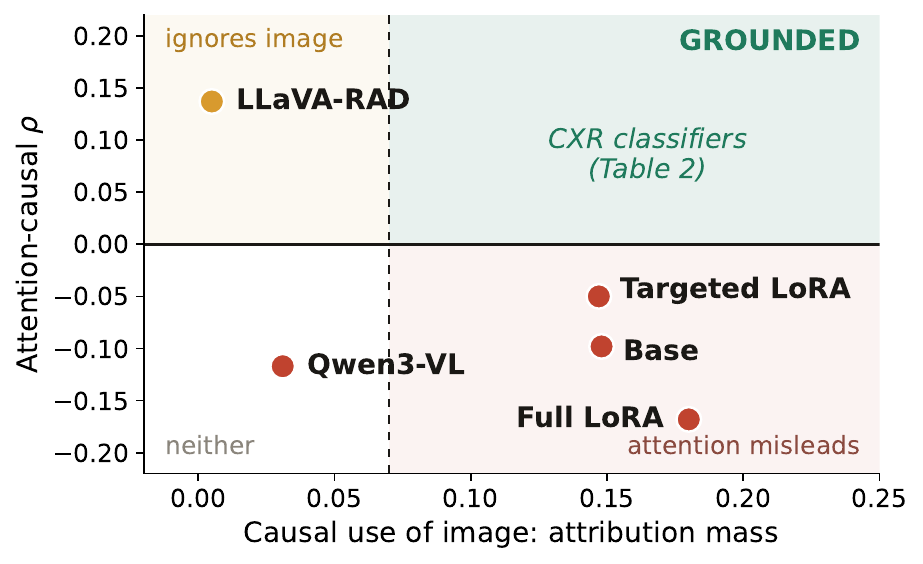}
\caption{Grounding needs both image use (horizontal: causal attribution mass) and attention-causal alignment (vertical: Spearman $\rho$). Points are exact Table~\ref{tab:counterfactual} values; the dashed line is a visual guide. No VLM reaches the grounded quadrant; the CXR classifier controls are grounded by Table~\ref{tab:attribution_methods}.}
\label{fig:quadrant}
\end{figure}

\textbf{No evaluated VLM has grounded attention.}
Grounding needs two conditions together (Fig.~\ref{fig:quadrant}): the MedGemma variants and Qwen3-VL use the image but their attention fails to track cause (negative Spearman $\rho$, surviving the $n{=}12$K multi-task scale-up), while LLaVA-RAD correlates positively only because it fails the first condition (causal attribution mass 0.005), so its $\rho$ joins two near-zero signals. CheXagent shows a decision can still be causally grounded in-distribution (VinDr), so we document that heatmaps do not certify grounding, not that medical VLMs never use the right region. The failure is not method- or modality-specific: no VLM method exceeds 22\% mask AMiM, and two CXR-trained classifiers pass the same protocol. Whether architectures with explicit grounding heads such as MIMO~\cite{mimo2025} fare better remains open.

\textbf{Scope and limitations.}
The DenseNet/ResNet contrast does not claim classifiers are globally better; it shows the protocol detects localized saliency when a model is trained for CXR classification. Causal maps use the $16{\times}16$ token grid, so sub-patch pathologies may be partly masked, and Qwen3-VL is evaluated at $8{\times}8$ for memory. A fill-sensitivity check yields no positive $\rho$ under any fill, so the reported value is a lower bound.

\textbf{Conclusion.}
Heatmaps from these medical VLMs are reassuring but not faithful: across bounding-box overlap, pixel-mask AMiM, and patch-occlusion Spearman $\rho$, attention fails where the CXR classifier baselines pass. Clinical explanations should be validated by controlled localization metrics and causal perturbation, not by visual inspection alone.

\noindent\textbf{Use of AI tools.} We used a large language model (Claude Opus 4.8) for language editing and LaTeX formatting (grammar, sentence-level rewriting, layout); it did not generate research questions, methods, experiments, or analysis. The authors verified all numbers, citations, and claims and take full responsibility.

\begin{credits}
\subsubsection{\ackname} Conducted at the Secure and Assured Intelligent Learning (SAIL) Lab, University of New Haven. No specific external funding was received.

\subsubsection{\discintname} B.~Sadanandan is a research engineer at Medtronic Medical Surgical, CT, USA; this employment is unrelated to the present work. The authors cite related primary studies of their own~\cite{anon_psfmed2026,anon_psflora2026,anon_uq2026}, published or under review elsewhere. No other competing interests are declared.
\end{credits}

\begingroup\scriptsize
\let\oldbib\thebibliography
\renewcommand{\thebibliography}[1]{\oldbib{#1}\scriptsize\setlength{\itemsep}{0pt}\setlength{\parsep}{0pt}\setlength{\topsep}{2pt}}

\endgroup

\end{document}